\documentclass[fleqn,10pt]{wlscirep}
\usepackage[utf8]{inputenc}
\usepackage[T1]{fontenc}
\usepackage{svg}
\usepackage{multirow}
\usepackage[super]{nth}
\usepackage{makecell}

\title{Predicting COVID-19 pandemic by spatio-temporal graph neural networks: A New Zealand's study}

\author[1,*]{Viet Bach Nguyen}
\author[2,*,+]{Truong Son Hy}
\author[3]{Long Tran-Thanh}
\author[4]{Nhung Nghiem}

\affil[+]{Correspondence to tshy@ucsd.edu}
\affil[*]{These authors contributed equally to this work.}

\affil[1]{FPT Software AI Center, Hanoi, Vietnam}
\affil[2]{University of California San Diego, La Jolla, United States}
\affil[3]{University of Warwick, Coventry, United Kingdom}
\affil[4]{University of Otago, Wellington, New Zealand}

\newtheorem{definition}{Definition}

\begin{abstract}
Modeling and simulations of pandemic dynamics play an essential role in understanding and addressing the spreading of highly infectious diseases such as COVID-19. In this work, we propose a novel deep learning architecture named Attention-based Multiresolution Graph Neural Networks (ATMGNN) that learns to combine the spatial graph information, i.e. geographical data, with the temporal information, i.e. timeseries data of number of COVID-19 cases, to predict the future dynamics of the pandemic. The key innovation is that our method can capture the multiscale structures of the spatial graph via a learning to cluster algorithm in a data-driven manner. This allows our architecture to learn to pick up either local or global signals of a pandemic, and model both the long-range spatial and temporal dependencies. Importantly, we collected and assembled a new dataset for New Zealand. We established a comprehensive benchmark of statistical methods, temporal architectures, graph neural networks along with our spatio-temporal model. We also incorporated socioeconomic cross-sectional data to further enhance our prediction. Our proposed model have shown highly robust predictions and outperformed all other baselines in various metrics for our new dataset of New Zealand along with existing datasets of England, France, Italy and Spain. For a future work, we plan to extend our work for real-time prediction and global scale. Our data and source code are publicly available at \url{https://github.com/HySonLab/pandemic_tgnn}.
\end{abstract}

\begin{document}

\flushbottom
\maketitle
%
%
\thispagestyle{empty}


\section{Introduction}
\label{ch:introduction}

The Coronavirus Disease started in 2019 (COVID-19) has been and currently is a major global pandemic, challenging every country's population and public health systems. As a fairly water-isolated island country, New Zealand mostly contained the spread of COVID-19 until early 2022, when infection cases surged to more than 2 million confirmed cases by the end of the year (WHO data, \url{https://covid19.who.int/region/wpro/country/nz}). While New Zealand responded promptly, contained and effectively vaccinated the population to keep the case number low, the sudden rise in infections posed certain challenges to the healthcare system. 

In the wake of the spread of COVID-19, many epidemiological modeling and prediction models emerged, seeking to project the progression of the pandemic and inform public health authorities to take measures when appropriate. Traditional dynamics case prediction models, including the family of Susceptible, Infectious, or Recovered models (SEIR) and their variants, have been widely applied to simulate the trajectories of the pandemic\cite{Wei2020-dx, Poonia2022-bc, doi:10.1080/03036758.2021.1876111}. SEIR models are hard to use due to difficulty in estimating parameters and underlying nonlinear systems of ordinary differential equations, alongside embedding restricted assumptions\cite{Roda2020-hg}. Other statistical prediction methods have been used, among which are the autoregressive integrated moving average (ARIMA) models\cite{RePEc:pes:ierequ:v:15:y:2020:i:2:p:181-204, ArunKumar2021-hu} and the time-series prediction Prophet model\cite{Kumar_2020}. However, linear statistical models cannot capture the non-linear nature of disease infection progression, especially in the case of New Zealand infection growth patterns. To model non-linear disease growth functions, artificial neural networks and deep learning models have been developed and trained to predict the infection case time series of each health area. The most common types of deep learning models for epidemic modeling are Long Short-Term Memory (LSTM)-based models, in which the architecture is specially designed to learn and represent historical or temporal information\cite{https://doi.org/10.48550/arxiv.1909.09586}. LSTM-based forecasting models can more accurately predict the number of cases and capture non-linear non-monotonous patterns in case data\cite{10.1371/journal.pone.0262708}, but otherwise cannot exploit crucial spatial or geographical information for predicting the spread of COVID-19 over multiple areas.

It is shown that incorporating geospatial information, including but not limited to movement and connectedness information, helps with the forecasting performance of LSTM-based and deep learning model\cite{cite-key-spatio}. One of the classes of deep learning models that can seamlessly embed geospatial information is Graph Neural Networks (GNNs), neural network deep learning models that can capture topological information in graph- and network-based data\cite{ZHOU202057}. Following in the footstep of previous efforts at COVID-19 forecasting with GNNs and spatial disease features\cite{Panagopoulos_Nikolentzos_Vazirgiannis_2021}, we propose improved spatiotemporal graph neural network models that can accurately learn and forecast COVID-19 case progression in New Zealand. To this end, we gathered and reformated New Zealand COVID-19 case data, and constructed day-to-day disease graphs based on geographical information; graph disease representations are then fed to a hierarchical, multi-resolution temporal graph neural network model that can automatically group multiple disease areas to learn large-scale disease properties\cite{https://doi.org/10.48550/arxiv.2106.00967, pmlr-v184-hy22a}. The \textit{Results} section \ref{ch:results} contains a demonstration of the graph-based New Zealand COVID-19 data format, various disease spread progression information, a performance comparison between different types of forecasting models, and an evaluation of the models' predictive capabilities. The \textit{Discussion} section \ref{ch:discussion} discusses the implications of the results on practical applications, as well as potential improvements and precautions. The \textit{Related works} section \ref{ch:related} presents a literature summary of the most relevant and up-to-date prior works related to our current work.The \textit{Methods} section \ref{ch:methods} provides details on the construction and mechanisms of the experimental epidemiological forecasting models.
\section{Results}
\label{ch:results}

\subsection{Descriptive data analysis}

\begin{figure}[htbp]
  \centering
  \includegraphics[scale=0.45]{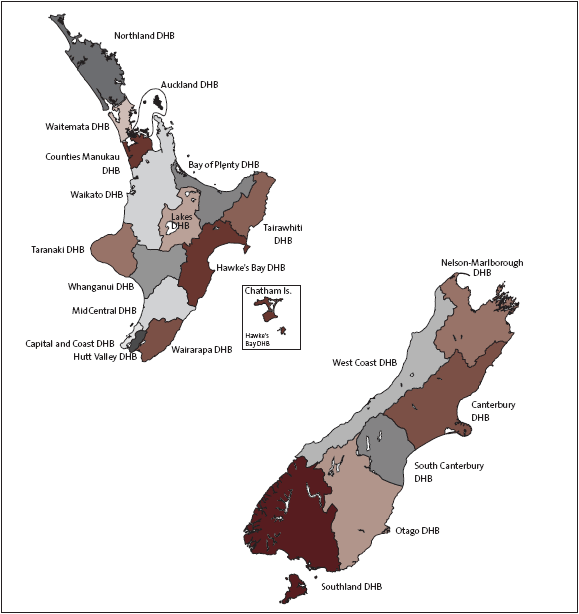}
  \caption{New Zealand's District Health Board map\cite{ministry_of_health_nz_2023}.}
  \label{fig:DHB_map}
\end{figure}

\begin{figure}[htbp]
  \centering
  \includegraphics[scale=0.55]{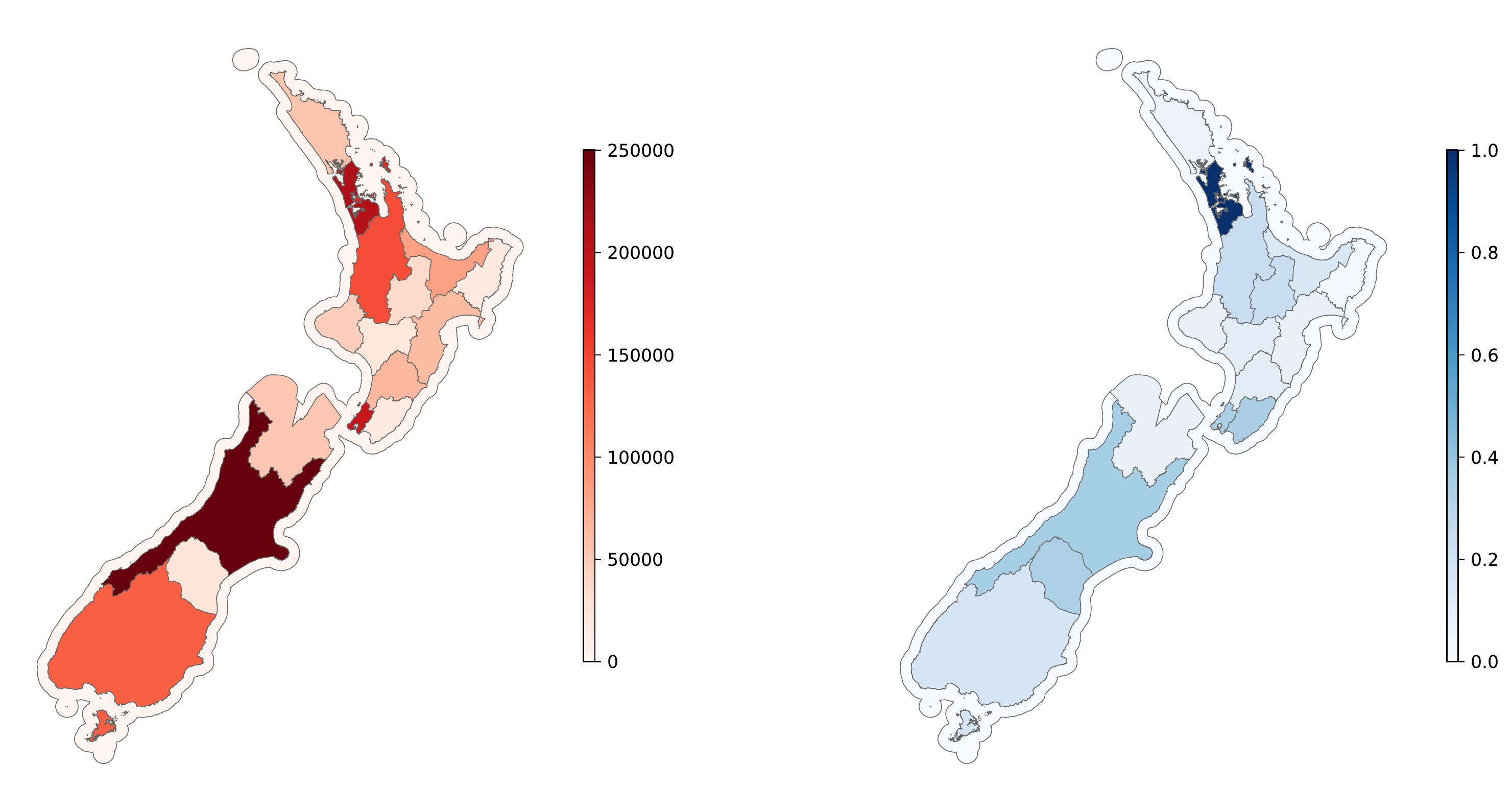}
  \caption{Left: The total number of COVID-19 cases by each district health board within the examination period. Right: GDP concentration across district health boards of New Zealand.}
\end{figure}

\begin{figure}[htbp]
  \centering
  \includegraphics[scale=0.65]{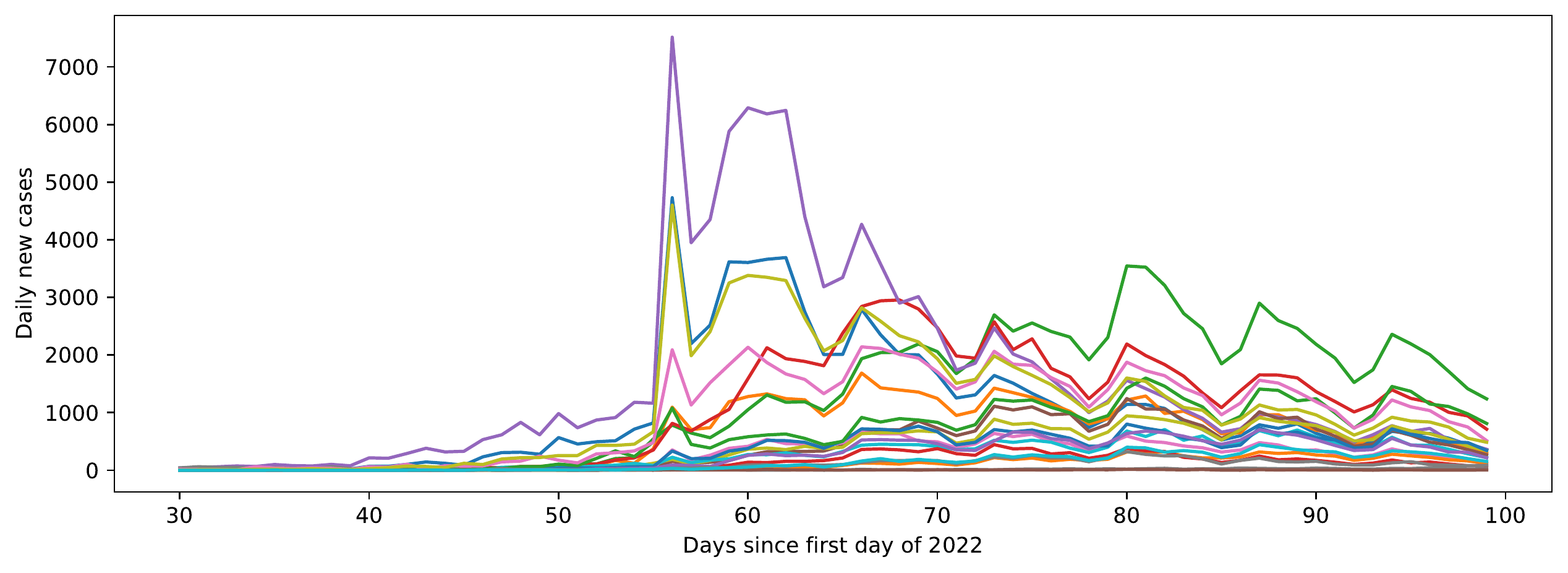}
  \caption{The number of daily new cases of all district health boards by the number of days since January \nth{1}, 2022.}
  \label{fig:nz_cases}
\end{figure}

\begin{figure}[htbp]
  \centering
  \includegraphics[scale=0.65]{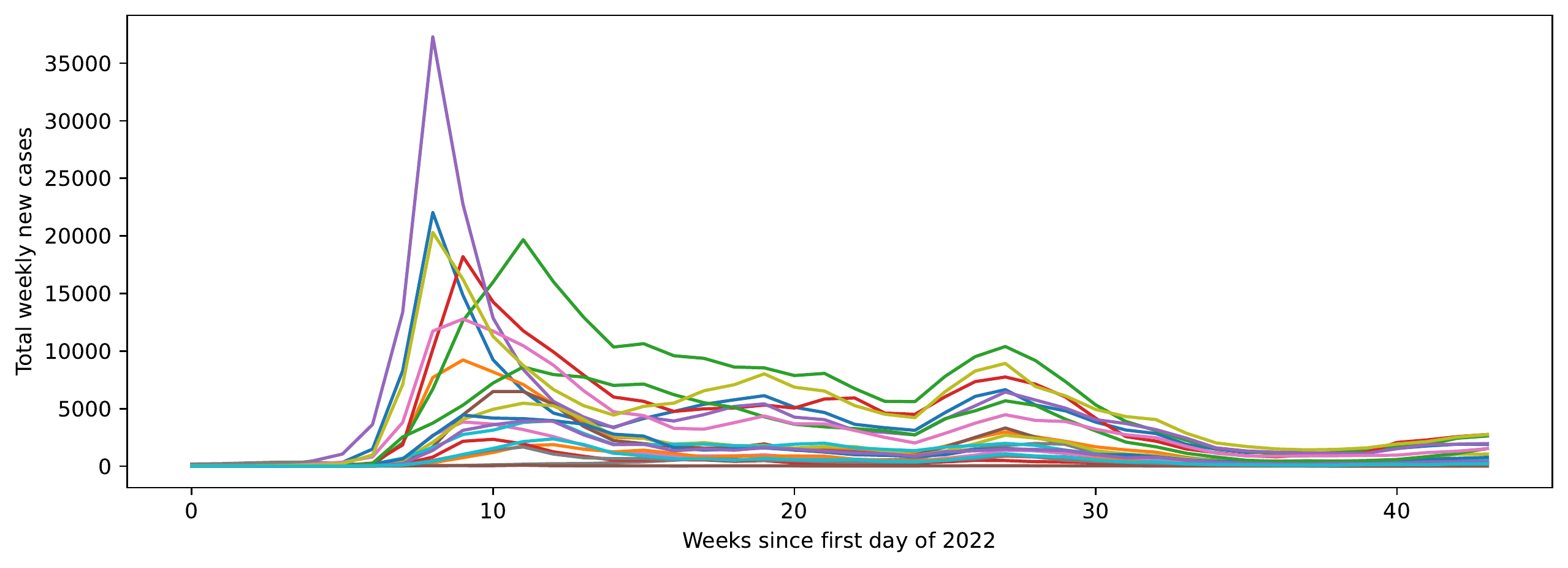}
  \caption{The number of total weekly new cases of all district health boards by the number of weeks since January \nth{1}, 2022.}
  \label{fig:nz_cases_week}
\end{figure}

The primary component of the dataset is the number of new cases in each region of the 20 district health boards in New Zealand (confirmed localized cases or general quarantine/unknown cases). Preliminary data analysis shows that there are significant variations between the different district health boards in terms of the volume of new cases quantified by mean and standard deviation, but the progression patterns are similar between different boards (Figure \ref{fig:nz_cases}). Spikes in new case counts are typically 5-7 days apart, indicative of either the disease incubation period\cite{Cheng2021} or spaced window flaw in data collection. Moreover, spikes between different regions generally coincide (Figure \ref{fig:nz_cases}), with the highest spikes and surges across the entire sampled time range occurring in March 2022. We confirmed that the case surges demonstrated by the dataset are consistent with existing pandemic reports during the same time frames\cite{Wolfe.2022}.

\subsection{Experiment task description}

\begin{table*}[t]
    \centering
    \def\arraystretch{1.1}
    \resizebox{\textwidth}{!}{ 
        \begin{tabular}{|l|rrr|rrr|rrr|rrr|} \hline
        \multirow{2}{*}{Model} & \multicolumn{3}{c|}{Next 3 Days} & \multicolumn{3}{c|}{Next 7 Days} & \multicolumn{3}{c|}{Next 14 Days} & \multicolumn{3}{c|}{Next 21 Days} \\ \cline{2-13}
        & MAE & RMSE & $\text{R}^2$ & MAE & RMSE & $\text{R}^2$ & MAE & RMSE & $\text{R}^2$ & MAE & RMSE & $\text{R}^2$ \\
        \Xhline{3\arrayrulewidth}
        AVG & 247.20 & 325.92 & -3.22 & 258.95 & 340.95 & -3.58 & 277.16 & 362.98 & -4.22 & 292.23 & 379.98 & -4.87 \\ 
        AVG\_WINDOW & 80.88 & 111.15 & 0.76 & 104.09 & 142.37 & 0.55 & 144.88 & 196.63 & -0.02 & 176.82 & 238.39 & -0.79 \\ 
        LAST\_DAY & 118.81 & 158.56 & 0.64 & 73.65 & 102.09 & 0.84 & 120.99 & 164.78 & 0.47 & 156.17 & 211.44 & -0.08 \\
        \hline
        LIN\_REG & 182.46 & 284.61 & 0.31 & 213.53 & 336.56 & -0.01 & 272.77 & 440.60 & -0.77 & 335.95 & 551.23 & -1.81 \\ 
        GP\_REG & 331.43 & 471.17 & -0.89 & 332.08 & 472.45 & -0.98 & 325.55 & 464.20 & -0.97 & 322.08 & 460.23 & -0.96 \\ 
        RAND\_FOREST & 98.97 & 152.96 & 0.80 & 72.85 & 111.69 & 0.89 & 112.02 & 168.81 & 0.74 & 140.77 & 210.73 & 0.59 \\ 
        XGBOOST & 109.68 & 165.36 & 0.77 & \textbf{68.51} & \textbf{105.91} & \textbf{0.90} & 108.45 & 165.17 & 0.75 & 137.45 & 208.13 & 0.60 \\ 
        \hline
        PROPHET & 119.32 & 642.78 & -0.24 & 148.58 & 770.55 & -1.50 & 222.01 & 526.58 & -0.59 & 292.54 & 407.66 & -0.17 \\ 
        ARIMA & 132.49 & 534.26 & 0.14 & 155.44 & 523.57 & -0.15 & 204.51 & 472.95 & -0.28 & 239.06 & 423.17 & -0.26 \\ 
        LSTM & 186.86 & 242.62 & -0.97 & 168.43 & 222.65 & -0.39 & 140.69 & 192.39 & 0.38 & 128.04 & 182.35 & 0.59 \\
        \hline
        \textbf{MPNN}  & 80.33 & 110.75 & 0.84 & 87.45 & 121.23 & 0.79 & 121.41 & 168.34 & 0.53 & 153.62 & 210.69 & 0.15 \\ 
        \textbf{MGNN}  & 80.87 & 111.67 & 0.83 & 89.77 & 124.56 & 0.74 & 125.30 & 172.46 & 0.46 & 156.25 & 213.55 & 0.06 \\ 
        \textbf{MPNN+LSTM} & \textbf{75.25} & \textbf{104.64} & \textbf{0.86} & 85.14 & 117.92 & 0.84 & \textbf{88.28} & \textbf{121.71} & \textbf{0.85} & \textbf{99.85} & \textbf{137.74} & \textbf{0.83} \\ 
        \textbf{ATMGNN} & 77.49 & 106.96 & \textbf{0.86} & 86.85 & 119.68 & 0.84 & 90.43 & 124.89 & 0.84 & 101.87 & 140.33 & 0.82 \\
        \hline
        \end{tabular}
    }
    \caption{New Zealand: Performance of all experimental model evaluated based on the metrics specified in Section \ref{sec: metrics}}
    \label{tab:NZ_results}
\end{table*}

\begin{table*}[t]
    \centering
    \def\arraystretch{1.1}
        \begin{tabular}{|l|rrr|} \hline
        \multirow{2}{*}{Model} & \multicolumn{3}{c|}{Linear decay slope}\\ \cline{2-4}
        & MAE & RMSE & $\text{R}^2$ \\
        \Xhline{3\arrayrulewidth}
        AVG & 2.551 & 3.069 & -0.090 \\ 
        AVG\_WINDOW & 5.451 & 7.265 & -0.081 \\ 
        LAST\_DAY & 4.194 & 5.743 & -0.067 \\ 
        \hline
        LIN\_REG & 8.457 & 14.739 & -0.112 \\ 
        GP\_REG & 2.491 & 3.539 & -0.028 \\ 
        RAND\_FOREST & 3.761 & 5.504 & -0.018 \\ 
        XGBOOST & 3.539 & 5.344 & -0.018 \\ 
        \hline
        PROPHET & 9.882 & \textit{-17.263} & \textit{0.025} \\ 
        ARIMA & 6.474 & -8.078 & -0.015 \\ 
        LSTM & \textit{-3.591} & \textit{-3.690} & \textit{0.107} \\
        \hline
        \textbf{MPNN}  & 4.651 & 6.339 & -0.044 \\ 
        \textbf{MGNN}  & 4.579 & 6.223 & -0.042 \\ 
        \textbf{MPNN+LSTM} & \textbf{1.134} & \textbf{1.511} & \textbf{-0.001} \\ 
        \textbf{ATMGNN} & 1.212 & 1.655 & -0.002 \\
        \hline
        \end{tabular}
    \caption{New Zealand: The slope of the linear fit to the performance decay graph of each model for all three metrics. For MAE and RMSE, lower slope values indicate better decay performance; higher $\text{R}^2$ slope values indicate better decay performance. The best-performing model is highlighted in \textbf{bold}, special exceptions are in \textit{italics}.}
    \label{tab:NZ_results_decay}
\end{table*}

\begin{table*}[t]
    \centering
    \def\arraystretch{1.1}
    \resizebox{\textwidth}{!}{ 
        \begin{tabular}{|l|rrr|rrr|rrr|rrr|} \hline
        \multirow{2}{*}{Model} & \multicolumn{3}{c|}{Next 3 Days} & \multicolumn{3}{c|}{Next 7 Days} & \multicolumn{3}{c|}{Next 14 Days} & \multicolumn{3}{c|}{Next 21 Days} \\ \cline{2-13}
        & MAE & RMSE & $\text{R}^2$ & MAE & RMSE & $\text{R}^2$ & MAE & RMSE & $\text{R}^2$ & MAE & RMSE & $\text{R}^2$ \\
        \Xhline{3\arrayrulewidth}
        AVG & 8.15 & 11.39 & -0.14 & 8.50 & 11.77 & -0.35 & 8.97 & 12.14 & -0.80 & 9.32 & 12.60 & -1.39 \\ 
        AVG\_WINDOW & 6.33 & 8.79 & 0.40 & 7.94 & 10.87 & -0.07 & 11.04 & 14.91 & -1.52 & 14.17 & 18.77 & -4.06 \\ 
        LAST\_DAY & 7.12 & 10.45 & 0.19 & 7.33 & 10.49 & 0.19 & 9.83 & 14.13 & -0.90 & 12.76 & 17.85 & -3.01 \\
        \hline
        LIN\_REG & 13.13 & 17.77 & -0.32 & 17.19 & 22.95 & -1.32 & 26.28 & 34.53 & -6.34 & 37.11 & 47.94 & -18.85 \\ 
        GP\_REG & 14.95 & 21.48 & -0.93 & 14.13 & 20.65 & -0.88 & 12.00 & 17.50 & -0.89 & 10.04 & 14.72 & -0.87 \\ 
        RAND\_FOREST & 6.64 & 9.87 & 0.59 & 7.16 & 10.17 & 0.54 & 10.01 & 14.08 & -0.22 & 13.03 & 17.88 & -1.76 \\ 
        XGBOOST & 7.32 & 10.93 & 0.50 & 7.46 & 10.82 & 0.48 & 10.00 & 14.50 & -0.30 & 12.82 & 18.06 & -1.82 \\ 
        \hline
        PROPHET & 10.79 & 20.78 & -0.06 & 14.45 & 29.28 & -1.29 & 23.43 & 34.29 & -2.99 & 33.59 & 31.72 & -3.66 \\ 
        ARIMA & 8.95 & 20.28 & -0.01 & 9.51 & 13.77 & 0.49 & 9.63 & 13.02 & 0.43 & 9.77 & 11.62 & 0.37 \\  
        LSTM & 8.61 & 11.88 & -0.36 & 8.20 & 11.24 & 0.12 & 7.86 & 10.66 & 0.47 & 7.09 & 9.95 & 0.65 \\
        \hline
        \textbf{MPNN}  & 6.51 & 9.41 & 0.55 & 7.54 & 10.63 & 0.39 & 10.12 & 14.14 & -0.42 & 12.84 & 17.77 & -1.82 \\ 
        \textbf{MGNN} & 6.87 & 9.48 & 0.55 & 8.18 & 10.93 & 0.35 & 11.07 & 14.67 & -0.60 & 14.14 & 18.67 & -2.16 \\ 
        \textbf{MPNN+LSTM} & 6.73 & 9.55 & 0.57 & 7.08 & 10.13 & 0.57 & 7.68 & 10.89 & 0.57 & 7.95 & 11.36 & 0.58 \\ 
        \textbf{ATMGNN} & \textbf{6.24} & \textbf{8.82} & \textbf{0.63} & \textbf{6.44} & \textbf{9.04} & \textbf{0.66} & \textbf{6.80} & \textbf{9.57} & \textbf{0.68} & \textbf{6.70} & \textbf{9.53} & \textbf{0.73} \\
        \hline        \end{tabular}
    }
    \caption{England: Performance of all experimental model evaluated based on the metrics specified in Section \ref{sec: metrics}}
    \label{tab:EN_results}
\end{table*}

\begin{table*}[t]
    \centering
    \def\arraystretch{1.1}
    \resizebox{\textwidth}{!}{ 
        \begin{tabular}{|l|rrr|rrr|rrr|rrr|} \hline
        \multirow{2}{*}{Model} & \multicolumn{3}{c|}{Next 3 Days} & \multicolumn{3}{c|}{Next 7 Days} & \multicolumn{3}{c|}{Next 14 Days} & \multicolumn{3}{c|}{Next 21 Days} \\ \cline{2-13}
        & MAE & RMSE & $\text{R}^2$ & MAE & RMSE & $\text{R}^2$ & MAE & RMSE & $\text{R}^2$ & MAE & RMSE & $\text{R}^2$ \\
        \Xhline{3\arrayrulewidth}
        AVG & 21.13 & 42.80 & 0.53 & 20.31 & 41.88 & 0.49 & 20.28 & 43.23 & 0.43 & \textbf{19.19} & 41.35 & 0.39 \\ 
        AVG\_WINDOW & \textbf{17.69} & \textbf{33.48} & \textbf{0.66} & 19.75 & \textbf{37.30} & 0.53 & 23.75 & 44.90 & 0.30 & 26.88 & 50.00 & -0.01 \\
        LAST\_DAY & 21.21 & 41.99 & 0.45 & 21.83 & 43.36 & 0.37 & 25.45 & 49.97 & 0.13 & 27.73 & 50.85 & 0.03 \\ 
        \hline
        LIN\_REG & 28.15 & 54.53 & 0.29 & 35.35 & 69.38 & -0.29 & 50.02 & 99.11 & -1.74 & 65.12 & 132.16 & -4.61 \\ 
        GP\_REG & 37.41 & 74.72 & -0.33 & 35.72 & 70.72 & -0.34 & 33.12 & 68.38 & -0.31 & 30.13 & 63.40 & -0.29 \\ 
        RAND\_FOREST & 19.42 & 40.80 & 0.60 & 20.41 & 42.38 & 0.52 & 24.80 & 50.01 & 0.30 & 27.41 & 52.76 & 0.11 \\ 
        XGBOOST & 21.79 & 47.09 & 0.47 & 22.49 & 48.63 & 0.37 & 26.41 & 55.60 & 0.14 & 28.64 & 57.06 & -0.05 \\
        \hline
        PROPHET & 23.03 & 55.65 & 0.46 & 29.18 & 71.44 & 0.15 & 40.95 & 92.95 & -0.76 & 51.92 & 100.06 & -1.64 \\ 
        ARIMA & 22.90 & 76.56 & -0.03 & 27.01 & 68.94 & 0.21 & 28.62 & 56.27 & 0.36 & 25.57 & 45.51 & 0.45 \\ 
        LSTM & 20.98 & 42.01 & 0.51 & 19.80 & 40.17 & 0.59 & \textbf{19.56} & 39.91 & 0.60 & 20.18 & 39.44 & 0.66 \\
        \hline
        \textbf{MPNN}  & 18.09 & 36.67 & 0.64 & 21.45 & 43.56 & 0.49 & 26.07 & 51.72 & 0.21 & 28.94 & 59.26 & -0.16 \\ 
        \textbf{MGNN} & 19.14 & 37.17 & 0.64 & 22.69 & 42.99 & 0.51 & 27.33 & 51.73 & 0.20 & 29.87 & 58.18 & -0.14 \\ 
        \textbf{MPNN+LSTM} & 18.50 & 38.43 & 0.60 & \textbf{19.48} & 39.98 & 0.59 & 19.72 & 41.89 & 0.56 & 19.84 & 41.22 & 0.58 \\ 
        \textbf{ATMGNN} & 18.05 & 36.94 & 0.65 & 19.63 & 39.06 & \textbf{0.63} & 19.80 & \textbf{39.49} & \textbf{0.63} & 18.55 & \textbf{37.07} & \textbf{0.67} \\
        \hline        \end{tabular}
    }
    \caption{Italy: Performance of all experimental model evaluated based on the metrics specified in Section \ref{sec: metrics}}
    \label{tab:IT_results}
\end{table*}

We comprehensively evaluate the forecasting effectiveness of the models in short-, mid-, and long-term prediction windows. The models are trained and assessed on their predictions 3, 7, 14, and 21 days from the input data. Data from day 1 to day $T$ is used to train one model at a time, and then predictions are obtained from the model from day $T+1$ to day $T+d$, where $d$ is the prediction window size and $0 < d < 22$. Note that each model within a single class of models is trained separately and specifically for a single fixed time window. In other words, two different models are trained to predict days $T + a$ and $T + b$, where $a, b>0$ and $a \neq b$. The size of the training set gradually increases as time progresses, and  for each value of $T$ the best model is identified via a validation set with no overlapping day with the test set. We trained and validated the models on the time series data from March \nth{4}, 2022 to September \nth{4}, 2022, and performed further model evaluations to examine generalization performance on an out-of-distribution starting from September \nth{4}, 2022 to November \nth{4}, 2022. Additionally, we ran experiments on all previous EU datasets 1-to-1 as mentioned in a previous study\cite{Panagopoulos_Nikolentzos_Vazirgiannis_2021} to further test the effectiveness of baseline models as well as more succinctly comparing graph spatiotemporal models to alternatives in a variety of disease settings.

\subsection{Baselines and Comparisons}

Several state-of-the-art COVID-19 forecasting models can be implemented as baseline models to measure the relative performance of our proposal models. Therefore, we compare the different spatio-temporal models with traditional statistical prediction and neural network-based regression models that have been recently applied to the problem of COVID-19 forecasting. Note that models that work with recovery, deaths, and policies such as SEIR are omitted since the dataset only provides the number of confirmed cases.

\paragraph{Simple statistical models} The class of most rudimentary statistical models for forecasting. The models examined include (1) AVG: The average number of cases for one region up to the time of the test day (e.g., the prediction for day 13 is based on the average number of cases of the last 12 days); (2) AVG\_WINDOW: The average number of cases in the past $d$-day window for one region (e.g., for $d=7$, the prediction for day 13 is based on the average number of cases of the last 7 days, from day 6 onwards); and (3) LAST\_DAY: The prediction for the next day in one region is the same as the number of cases on the previous day.

\paragraph{Traditional machine learning models} The input format for all models in this class is the case history up to the prediction date of each district health board. The models examined include (4) LIN\_REG\cite{10.1007/978-981-19-6004-8_17}: Ordinary least squares Linear Regression, which fit a line onto seen training samples to predict the number of future cases (linear approximation); (5) GP\_REG\cite{Ketu2021-sq}: Gaussian Process Regressor, a non-parametric based regression model commonly used in predictive analysis that implements Gaussian processes; (6) RAND\_FOREST\cite{Galasso2022-eg}: A random forest regression model that produces case predictions using decision trees, with multiple trees built based on training case data to best average the final results; and (7) XGBOOST\cite{Fang2022-gf}: An improved version of the random forest regression model using gradient boosting.

\paragraph{Parameterized regression time-series forecasting models} The class of linear regression models with specific components represented as parameters. The models examined include (8) PROPHET\cite{Mahmud2020BangladeshCD}: A forecasting model for various types of time series that has also seen extensive use in forecasting COVID-19 where the input is the entire time-series historical number of cases of one region up to before the testing day; (9) ARIMA\cite{RePEc:pes:ierequ:v:15:y:2020:i:2:p:181-204}: A simple autoregressive moving average model, which the input is similar to PROPHET; and (10) LSTM\cite{10.1371/journal.pone.0262708}: A two-layer bidirectional LSTM model that takes as input the sequence of new cases in a region for the last 7 days, popular for the forecasting task capable of state-of-the-art performance.

\paragraph{Graph neural network-based models} The proposal graph models to be compared to the previous baseline models, with and without temporal components. The models examined include (11) MPNN\cite{pmlr-v70-gilmer17a}: Message-passing neural network model with separate layers for each day in the case time series; (12) MGNN\cite{pmlr-v184-hy22a, Hy_2023}: Message-passing neural network model similar to MPNN, but with multiple graph resolutions and learned clustering of different regions; (13) MPNN + LSTM: Message-passing neural network model with long-short term memory neural time series model; and (14) ATMGNN: Multiresolution graph model based on the MGNN model combined with Transformers for modeling time series. All models in this category are described in detail in Section \ref{ch:methods}.

\subsection{Experimental setup}

We detail the hyperparameters setup of the deep learning prediction models in our experiments. For all graph-based models (MPNN, MPNN+LSTM, ATMGNN), training lasts for a maximum of 300 epochs with early stopping patience of 50 epochs, and early stopping is only enabled from epoch \nth{100} onward. Models are trained with the Adam optimizer ($lr=10^{-3}$), batch size 128. For the neighborhood aggregation layers of the graph models, batch normalization is applied to the output of all layers with dropout applied to 0.5 times the total number of nodes. The LSTM component of the MPNN+LSTM model is implemented with a hidden state size of 64. The multiresolution component of the ATMGNN is configured for two additional coarsening layers with 10- and 5-node clusters, respectively; self-attention is configured with a single head for all regions. The models with the lowest validation loss at each prediction day shift are saved as parameter checkpoints for the sake of further evaluation, and validation information is outputted for further examination. All models are implemented with PyTorch\cite{NEURIPS2019_bdbca288} and PyTorch geometric\cite{Fey/Lenssen/2019}.

\subsection{Evaluation metrics}
\label{sec: metrics}

We measured the performance of all models with three evaluation metrics: Mean absolute error (MAE), root mean squared error (RMSE), and the coefficient of determination ($\text{R}^2$-score).

\paragraph{Mean Absolute Error (MAE)} The metric measures the sum of the absolute differences between the predicted case count and the actual case count. MAE cannot indicate the degree of under-/over-prediction since all variations have equal weight. The formula to calculate MAE is as follows:
\begin{equation*}
    \text{MAE} = \frac{1}{N}\sum^N_{i=1}|\hat{y}_i - y_i| 
\label{eq:MAE}
\end{equation*}
where $\hat{y}_i$ represents the forecast case count, $y_i$ represents the ground truth case count, $N$ represents the total number of days in the case count time series, and $i$ represents the case count statistic of a single day in the time series.

\paragraph{Root Mean Squared Error (RMSE)} The metric measures the square root of the average squared deviation between the forecast and the actual case count. RMSE is a good measure of prediction accuracy, mostly used when the error is highly nonlinear. The formula to calculate RMSE is as follows:
\begin{equation*}
    \text{RMSE} = \sqrt{\sum^N_{i=1}\frac{(\hat{y}_i - y_i)^2}{N}}
\label{eq:RMSE}
\end{equation*}

\paragraph{Coefficient of Determination ($\text{R}^2$)} The metric represents the proportion of variance of the case count that has been explained by the independent variables in the model. $\text{R}^2$ indicates goodness of fit and measures how well unseen samples are likely to be predicted by the model (through the proportion of explained variance). The formula to calculate $\text{R}^2$-score is as follows:
\begin{equation*}
    \text{R}^2 = 1 - \frac{\sum^N_{i=1}(\hat{y} - y_i)^2}{\sum^N_{i=1}(\bar{y} - y_i)^2}
\label{eq:R2-score}
\end{equation*}
where $\bar{y} = \frac{1}{N}\sum^N_{i=1}y_i$, or the average of the ground truth time series. The best possible score is 1.0, and the score can be negative (i.e., the model can arbitrarily badly fit the case count time series). A constant model that always predicts the average number of cases over the entire periods with no respect to the inputs would get an $\text{R}^2$ score of 0.0.

\subsection{Observations}

\paragraph{Performance measurement} Table \ref{tab:NZ_results} details the performances of all experimental models in the benchmark study. Across the board, MPNN+LSTM is the highest-performing model, with relatively low mean error and root means square error, alongside accurate trend prediction at $R^2$-score consistently over 0.8. Other baseline methods performed inconsistently across different time ranges, with massive fluctuations in heuristic statistical methods (AVG, AVG\_WINDOW, and especially LAST\_DAY), owning to these baselines simply forecasting based on rudimentary statistics of the data. However, in near-future time prediction windows (from 1-7 days), simple statistical methods can be competitive compared to more complex models; nevertheless, our goal is to eliminate or mitigate performance decay during long-term predictions. The class of traditional machine learning models performed reasonably well, with tree-based methods RAND\_FOREST and XGBOOST outperforming simple statistical methods and parameterized models aside from LSTM. In New Zealand from 14 days prediction length onwards, graph-based temporal models on average see a 20.31\% and 25.37\% relative reduction in MAE and RMSE, respectively; while correlation metric $\text{R}^2$ relatively improving 9.43\%. Similar results are obtained from cross-examining Italy and England COVID datasets, with graph-based temporal models (e.g., MPNN+LSTM, ATMGNN) generally outperforming other baseline models and LSTM models coming second on all metrics. The exception to performance patterns across countries is the overall performance of traditional machine learning models, which perform inconsistently across different countries. France and Spain's tables are included in the Appendix.

\begin{figure}[htbp]
  \centering
  \includegraphics[scale=0.7]{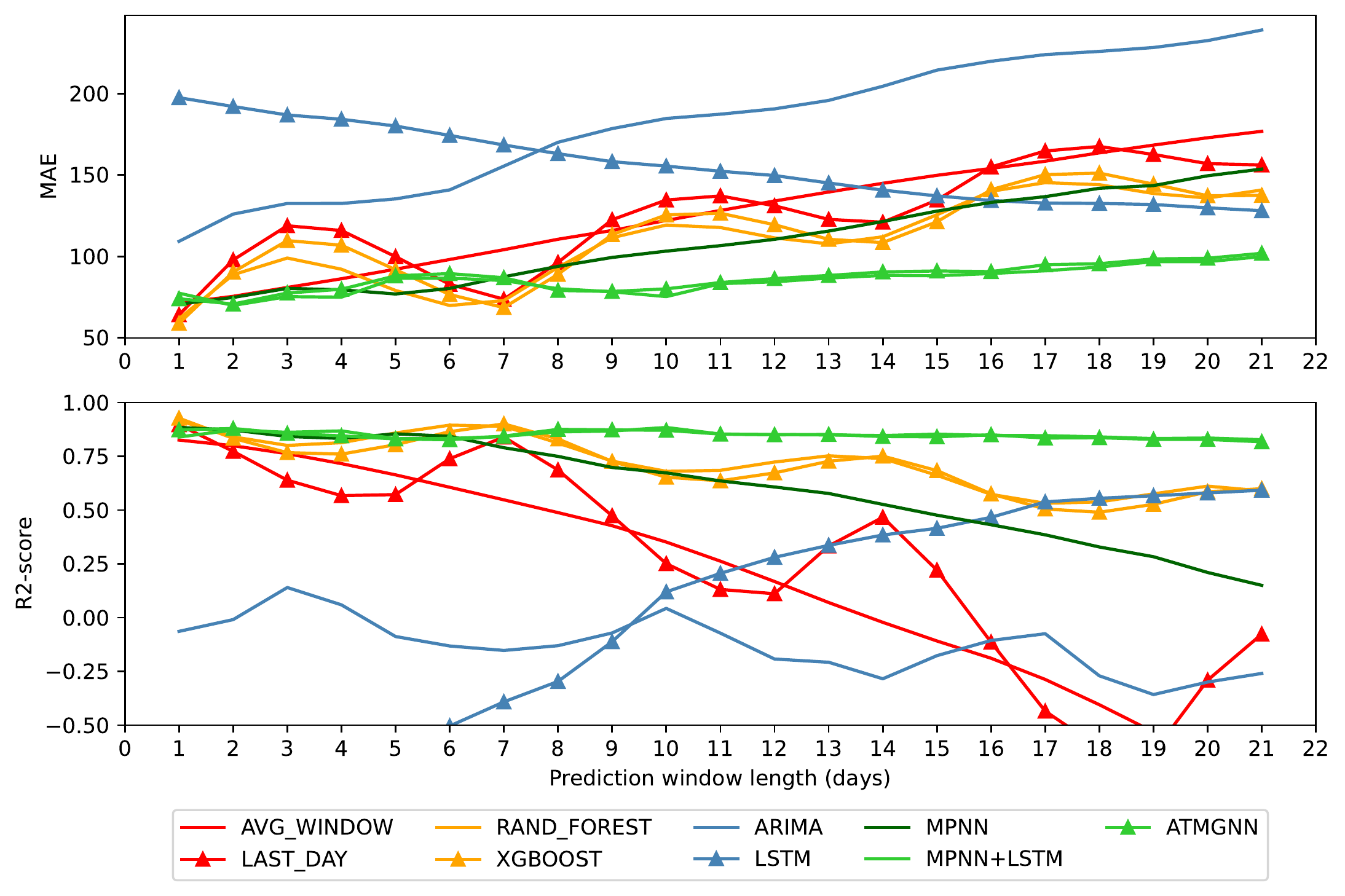}
  \caption{Performance decay with respect to MAE and $\text{R}^2$ metrics. Models with performance worse than the defined y-axis range are excluded.}
  \label{fig:NZ_decay}
\end{figure}

\begin{figure}[htbp]
  \centering
  \includegraphics[scale=0.7]{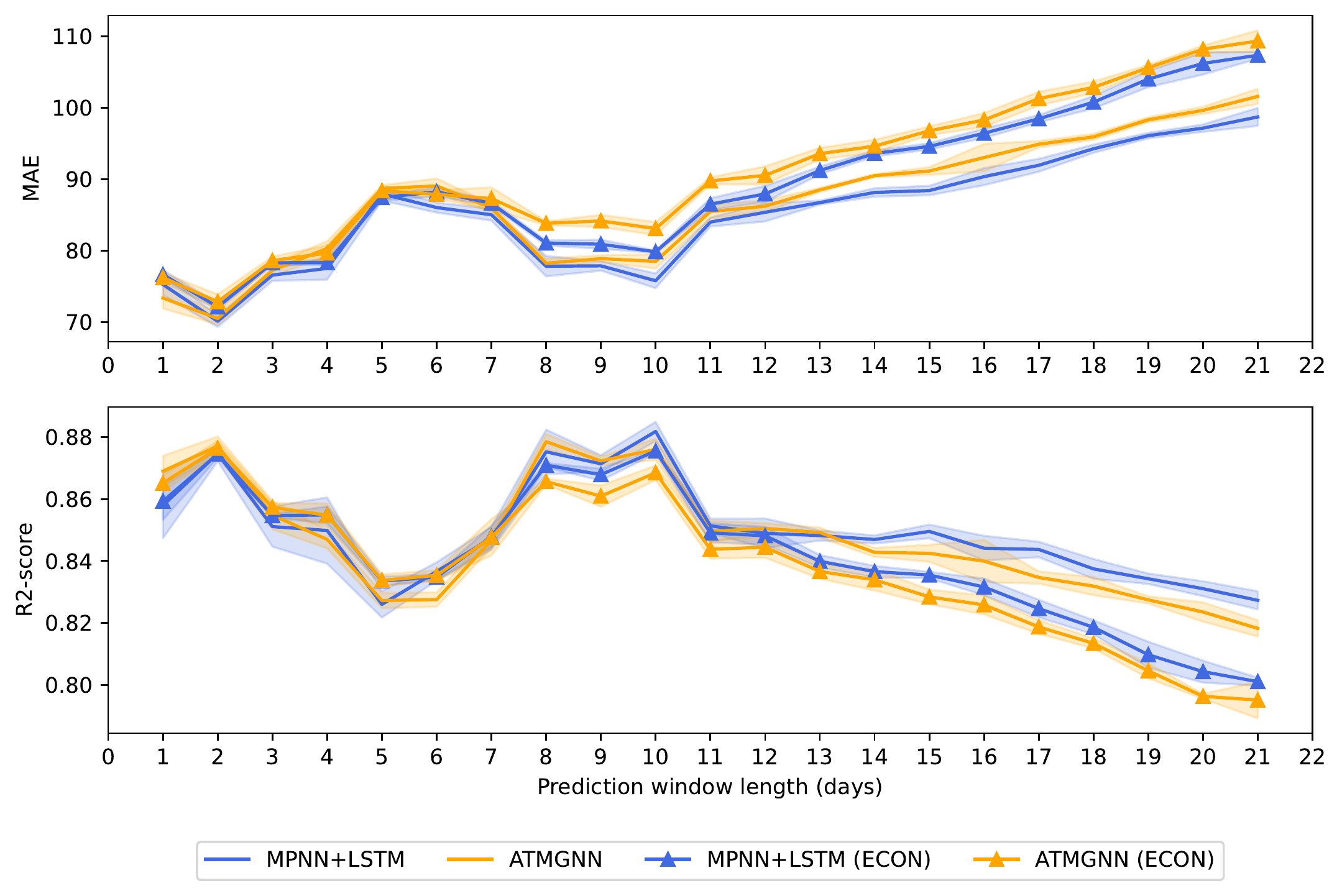}
  \caption{Average error and R2-score decay of two temporal graph network models over multiple runs.}
  \label{figure:tgnn}
\end{figure}

\paragraph{Performance decay over long forecasting windows} Across all models, the AVG model performed the worst when it comes to performance decay relative to the length of forecasting windows, concerning both absolute error and correlation metrics. On the other hand, the two other heuristic statistical methods, AVG\_WINDOW and LAST\_DAY, outperformed regression-based methods ARIMA and PROPHET with respect to the rate of decay and error increment over longer forecasting windows (Figure \ref{fig:NZ_decay} and Table \ref{tab:NZ_results_decay}). We observed anomalies in the results of our LSTM implementation: both the model's absolute error and correlation score $R^2$ do not decay over time, but rather improved over the long run (Figure \ref{fig:NZ_decay}); this phenomenon may be explained considering LSTM-based models performed better in certain long time ranges, or the models have certain "sweet spots". Traditional machine learning models that implement tree-based learning, including RAND\_FOREST and XGBOOST, performed reasonably well in terms of performance decay, with lower error increase rate and $\text{R}^2$ decrease rate than most other models aside from graph-based temporal models and ARIMA. Graph- and temporal-hybrid models MPNN+LSTM and ATMGNN maintained a stable performance decay profile with a low decay rate on both error and correlation compared to every other model aside from the LSTM exception, alongside lower values in both metrics across the board. Both temporal graph models started with relatively high performance in terms of all metrics compared to other baselines and mostly maintained the same performance when predicting longer time ranges with minimal decay, resulting in them outperforming all other baseline models. We specifically demonstrated the relative metric and decay stability of the two graph-based temporal models by averaging over several runs and computing the deviation as in Figure \ref{figure:tgnn}, showing the performance and decay similarities between these models over time.

\begin{figure}[htbp]
  \centering
  \includegraphics[scale=0.55]{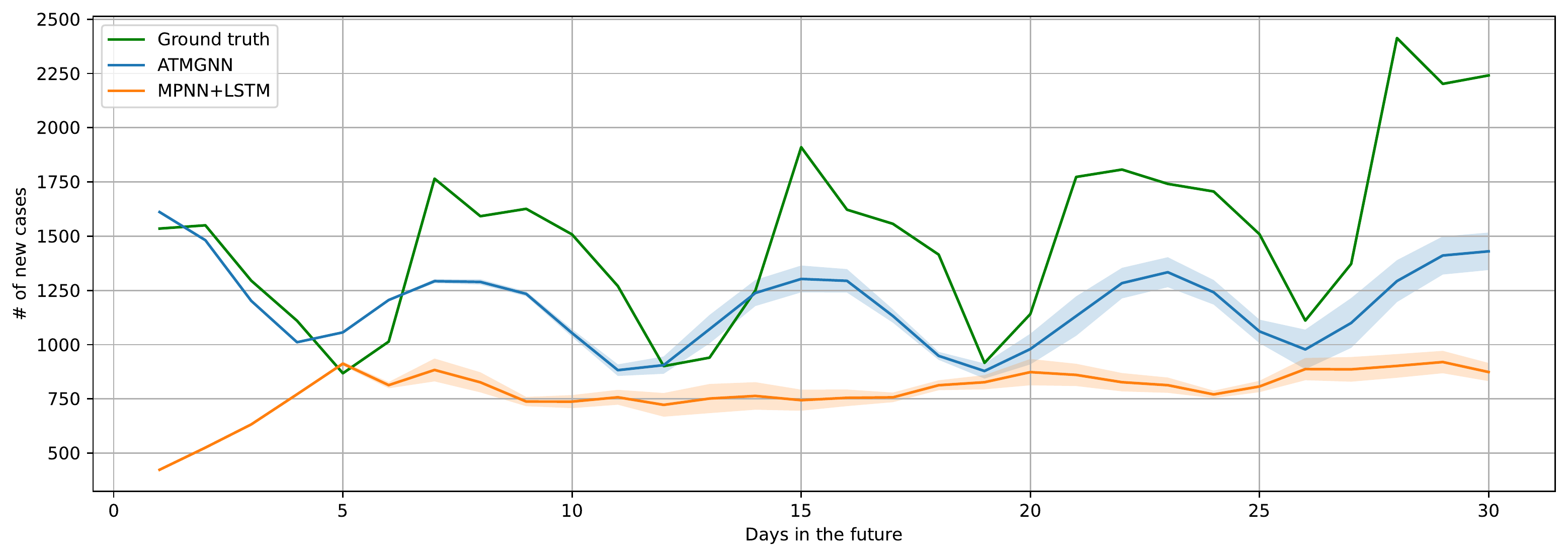}
  \caption{Sample out-of-distribution new case predictions for two graph-based models.}
  \label{fig:oop}
\end{figure}

\paragraph{Out-of-distribution forecasting} We examined the out-of-distribution performances of two of our best-performing models, the MPNN+LSTM and ATMGNN. The evaluation is done on the number of new cases between September \nth{4}, 2022 and November \nth{4}, 2022, with no overlapping between the evaluation set and the train/validation sets. All models are evaluated as autoregressive models, meaning for the 30-day prediction window the models use the prediction output of the previous day as an input feature for predicting the current day. As demonstrated in Figure \ref{fig:oop}, ATMGNN outperformed MPNN+LSTM when it comes to prediction error and emulating the spiking dynamics of the number of new cases. The predictions retrieved from the outputs of ATMGNN showed that the model can fairly precisely simulate the case spiking dynamics even when tested on a dataset completely unrelated and separate from the training dataset, demonstrating good generalization performance of the model. All other baseline models are not included in the evaluation after extensive testing showing that their performances are not remotely comparable to the two models demonstrated above. Further testing with different information windows with ground truth case information feed into the model ranging between 3 and 9 days before the target day to predict showed that both models maintained similar forecasting patterns with ATMGNN better conforming to the ground truth and the models maintaining relatively stable predictions given different case information levels. 

\begin{table*}[t]
    \centering
    \def\arraystretch{1.1}
    \resizebox{\textwidth}{!}{ 
        \begin{tabular}{|l|rrr|rrr|rrr|rrr|} \hline
        \multirow{2}{*}{Model} & \multicolumn{3}{c|}{Next 3 Days} & \multicolumn{3}{c|}{Next 7 Days} & \multicolumn{3}{c|}{Next 14 Days} & \multicolumn{3}{c|}{Next 21 Days} \\ \cline{2-13}
        & MAE & RMSE & $\text{R}^2$ & MAE & RMSE & $\text{R}^2$ & MAE & RMSE & $\text{R}^2$ & MAE & RMSE & $\text{R}^2$ \\
        \Xhline{3\arrayrulewidth}
        MPNN+LSTM & 75.25 & 104.64 & 0.86 & 85.14 & 117.92 & 0.84 & 88.28 & 121.71 & 0.85 & 99.85 & 137.74 & 0.83 \\ 
        ATMGNN & 77.49 & 106.96 & 0.86 & 86.85 & 119.68 & 0.84 & 90.43 & 124.89 & 0.84 & 101.87 & 140.33 & 0.82 \\
        \hline 
        \textbf{MPNN+LSTM (ECON)} & 78.30 & 108.11 & 0.85 & 86.63 & 119.16 & \textbf{0.85} & 93.61 & 129.13 & 0.84 & 107.37 & 147.57 & 0.80 \\ 
        \textbf{ATMGNN (ECON)} & 78.61 & 108.54 & 0.86 & 87.30 & 120.23 & \textbf{0.85} & 94.64 & 130.53 & 0.83 & 109.36 & 150.02 & 0.80 \\
        \hline
        \end{tabular}
    }
    \caption{New Zealand: Performance measurements of graph-based models with and without economic features}
    \label{tab:NZ_results_econ}
\end{table*}

\paragraph{Economic graph features} To further analyze the role of auxiliary features in graph-based models, we integrated economic features into the two best-performing graph-based forecasting models. Details on the source and construction of these features within the dataset are discussed in Section \ref{sec:DP}. As shown in Figure \ref{figure:tgnn} and Table \ref{tab:NZ_results_econ}, the two models that include economic features (ECON models) slightly underperformed compared to baseline graph-based temporal models, even with all economic features normalized. The decrease in performance is likely due to these economic features remaining constant throughout all prediction periods, indicating that relative economic allocation between different district health boards do not necessarily add any helpful information. Because of either low baseline performance or incompatibility, such as the economic zones and DHBs were not strictly overlapped as per Figure \ref{fig:DHB_map}, these economic features were not added to other baseline models. Our sensitivity analysis that removed DHBs with inconsistent economic zones suggested there was a small improvement in model performance as expected (see the Appendix).

\paragraph{Demographic graph features} We additionally tested different modalities of the original graph models, particularly enhancing the input to the model with separate age group data from the original data source, as well as augmenting the output of the models to predict the number of new cases for each age group of each district health board in New Zealand. Through testing, however, outputting each age group separately, adding age group data to the models' inputs, as well as adding custom demographic weighting to each age group during training did not improve the models' performance. For reference, the results of testing demographic-enhanced models are presented in the Appendix.

\section{Discussion}
\label{ch:discussion}
 
\paragraph{Interpretation of the main results} We provided a comprehensive evaluation of four classes of COVID-19 forecasting models, with a detailed analysis of the models' performance, decay over time, out-of-distribution forecasting, and economic features addition. Generally, graph neural network-based models, specifically the temporal variant of graph-based models outperformed every other baseline model in terms of performance metrics and performance decay over time. This trend is not shared by non-temporal graph-based models, indicating the importance of temporal mechanisms in forecasting models, whether it is attention-based on recurrent network-based (i.e., LSTM). Additionally, for far less computation cost, traditional machine learning models outperformed statistical and parameterized regression time-series forecasting models, even remaining competitive with the neural network-based LSTM model. The results suggest that the spatiotemporal approach to modeling the spread of COVID-19 based on the number of new cases is effective compared to other traditional modeling methods. Intuitively, graph-based models can accurately simulate the change in the number of new cases in one region when given that region's traffic connectivity with its neighbors. Since the spread of COVID-19 in every country, not only in New Zealand, is movement-based in nature, by modeling such geographical connectivity we can find latent information by accounting for human contacts with graph-based models. Moreover, the out-of-distribution performance of multiresolution temporal graph models also demonstrates the utility of modeling the problem of COVID-19 forecasting as a hierarchical system, with spreads localized in adjacent regions that have significant traffic volume.

\paragraph{Strength} Graph-based temporal models can embody the hidden correlations between different district health board regions when forecasting COVID-19. The approach is straightforward with versatility in modeling various connected systems and structures, not only with the task of forecasting disease spread trajectories but also in the task of chemical molecule construction and discrimination\cite{xia2023molebert}. As indicated by the performance metrics, temporal models perform well compared to other traditional models, predicting with relatively low errors and remaining accurate even when predicting further into the future. The multiresolution setting of graph-based models can also accurately model the disease's case dynamics to a reasonable degree when facing completely new and separate data from the training set. Additionally, the models' current implementation relies on comparatively low computational resources, with training and inferencing solely based on an online connection to an instance of Google Colab. Our COVID-19 data were captured during the Omicron waves in New Zealand with a highly vaccinated population. Our data also reflected a unique social experimental event that the New Zealand border was opened in May 2022 after strictly closing for two years (since 25 March 2020).

\paragraph{Limitations} While certain metrics of the proposed models are satisfactory, we have identified several weaknesses of the models that were tested. Graph-based models, while powerful, still require a certain amount of computational resources and adequate time for the process of training the models. Data inputs also have to be well-structured and preprocessed carefully to suit the formatting of the models, though this is less of a concern given the availability and accuracy of case datasets such as the New Zealand COVID-19 public dataset. Furthermore, data features can be further enriched with more detailed movement data between regions, traffic density information for all traveling modalities (e.g., land, sea, or air travel), and local movement details within each region. Most importantly, graph-based networks and deep learning models in general are black-boxes, offering little insight into the precise mechanisms of forecasting and modeling disease dynamics for the sake of studying the exact nature of epidemic spread.

\paragraph{Policy implications} The out-of-distribution performance and time range availability of up to 30 days show that graph-based multiresolution temporal models can be effectively used in aiding public health policies. With appropriate data processing and extension, the models can be adapted to predicting new cases on coarser temporal resolutions (i.e., weeks or months), potentially becoming a useful tool for epidemic predictive modeling and local or country-level intervention measures simulations.

\paragraph{Future research} Future directions of research are aplenty, from additional data features and enrichment features readily incorporable into the model as node features (e.g., more fine-grained socioeconomic features) or as edge features (e.g., mobility data), to interpretation methods designed for graph neural networks\cite{li2022survey} for the sake of understanding the inner workings of such prediction models. Another interesting direction is to comparatively examine the spatial modeling of deep learning models with other dynamical forecasting models, and the correlation influence of each disease feature/parameter on the final prediction output of each type of model.

\paragraph{Conclusions} Our study suggested that graph neural network-based models outperformed every other baseline model in terms of performance metrics and performance decay over time. Furthermore, our graph neural network-based models can effectively predict the mumber of COVID-19 cases upto 30 days, and therefore can assist with public health policies planning in order to control the COVID-19 outbreaks. Finally, our results in terms of model structures and frameworks can be generalized to other countries with similar settings.
\section{Related works}
\label{ch:related}

\subsection{Linear and statistical forecasting models}

Various traditional statistical and linear models have been employed to forecast the spread of COVID-19 cases. Among these traditional models are the Susceptible, Infectious, or Recovered (SEIR) models, where the dynamics of disease spread is modeled as a function of various population and the interactions between them\cite{Wei2020-dx, Poonia2022-bc, doi:10.1080/03036758.2021.1876111}. While mathematical models such as SEIR can estimate the effect of control measures even before the start of the pandemic, these models cannot make accurate predictions due to a lack of data and their inherent assumptions restricting the class of available learnable disease functions\cite{Roda2020-hg, CuestaHerrera2022AnalysisOS}.

Some other prevalent classes of forecasting methods are the Autoregressive Integrated Moving Average models (ARIMA) and the time-series prediction Prophet model. ARIMA models are well-known for being able to forecast future points in time series data, especially when the mean of the data is non-stationary; evidently, this family of models has been applied numerous times to forecast COVID-19 cases in several countries\cite{RePEc:pes:ierequ:v:15:y:2020:i:2:p:181-204, ArunKumar2021-hu, YOUSAF2020109926}. On the other hand, the Meta-developed Prophet model and its variants have also been utilized to tackle the task of predicting the progression of COVID-19 cases with some success, namely in forecasting the number of cases in India\cite{ILKOM1219} and generally for any country using day level case information\cite{Kumar_2020, Battineni2020ForecastingOC}. Several combinations of the traditional and statistical linear models have been examined\cite{doi:10.1080/09720502.2020.1833458}, effectively achieving better performance using compositions of successful statistical time-series models.

While statistical models are proficient at capturing certain COVID-19 case dynamics and are similarly motivated by repeated case patterns shown in pandemic data, these models are linear in nature and incapable of modeling spatial information as well as higher-level disease functions. Among the above models, ARIMA-based models have been proven to be insufficient even in their specialty of modeling time-series data given the complexities of such type of data in certain dimensions, while also underperformed compared to simple empirical methods\cite{https://doi.org/10.48550/arxiv.1904.07632, HYNDMAN20207}.

\subsection{Neural networks-based time-series forecasting models}

Neural networks are a type of machine learning method that is capable of learning arbitrary non-linear functions underlying the data, making them one of the most powerful general learning algorithms for a wide range of tasks\cite{Hornik1989MultilayerFN}. The vanilla neural networks without any additional component have been tested as predictors of COVID-19 outbreaks across several countries, owing to their high capacity modeling of disease patterns and functions when certain assumptions (e.g., disease incubation period) are encoded\cite{Niazkar2020ApplicationOA}.

Even among powerful Artificial Neural Networks (ANNs), there arises a need to explicitly model the temporal nature of certain types of data (i.e., the time-dependent trends of disease outbreaks)\cite{Ismail_Fawaz_2019}. The most common methods for modeling time-series data, particularly pandemic forecasting input data, are Recurrent Neural Network (RNN) models. RNNs are widely adopted in areas with sequential data; intuitively, RNNs are capable of "memorizing" the nature of patterns within time-series data. Most popular within the class of RNNs are Long-Short Term Memory (LSTM) models, a variant of the traditional RNNs that is capable of modeling long-term dependencies, allowing the models to learn information and patterns from distant past\cite{8737887}. LSTMs and their variants have been used extensively to forecast COVID-19 case progression, notably in Canada, where the model was tested and modified to accommodate disease-specific information, accurately predicting an exponential surge in the number of cases\cite{Chimmula2020TimeSF}. LSTM-based models were also used to simulate and forecast the COVID-19 pandemic in several other countries, either independently or in conjunction with various distinct statistical models incorporating spatial features\cite{Nikparvar2021SpatiotemporalPO, Luo2021TimeSP, Shastri2020TimeSF}.

While time-series models are effective at modeling the evolution of the disease over time in a single specific geographical region, LSTM/RNN-based models inherently cannot incorporate spatial features without employing heuristics that either explicitly change the architecture or interpolate spatial information into the input data itself\cite{Organero2022DeepSM}. To address the shifting topology of the natural geographical map and represent the causal relationship between pandemic regions, a more graphical and hierarchical approach is needed.


\subsection{Temporal graph neural networks forecasting models}

Graph neural networks (GNNs) utilizing various ways of generalizing the concept of convolution to graphs \cite{Scarselli09,Niepert2016,Li2016} have been widely applied to many learning tasks, including modeling physical systems\cite{NIPS2016_3147da8a}, finding molecular representations to estimate quantum chemical computation\cite{Duvenaud2015,Kearnes16,pmlr-v70-gilmer17a,HyEtAl2018,hy2019covariant}, and protein interface prediction\cite{10.5555/3295222.3295399}. One of the most popular types of GNNs is message passing neural nets (MPNNs)\cite{pmlr-v70-gilmer17a} that are constructed based on the message passing scheme in which each node propagates and aggregates information, encoded by vectorized messages, to and from its local neighborhood. In order to capture the dynamic nature of evolving features or connectivity over time, temporal graph neural networks (TGNN) have been proposed by \cite{rossi2020temporal,pmlr-v184-hy22a} as a generic, efficient deep learning framework that combines graph encoding (e.g., MPNNs) with time-series encoding architectures (e.g., LSTM, Tranformers, etc.). Applications of TGNN include traffic prediction\cite{Diao_Wang_Zhang_Liu_Xie_He_2019,li2018diffusion,nguyen2023fast} and learning on brain networks\cite{nguyen2023fast}, etc.
\section{Methods}
\label{ch:methods}



\begin{figure}[htbp]
  \centering
  \includegraphics[scale=0.55]{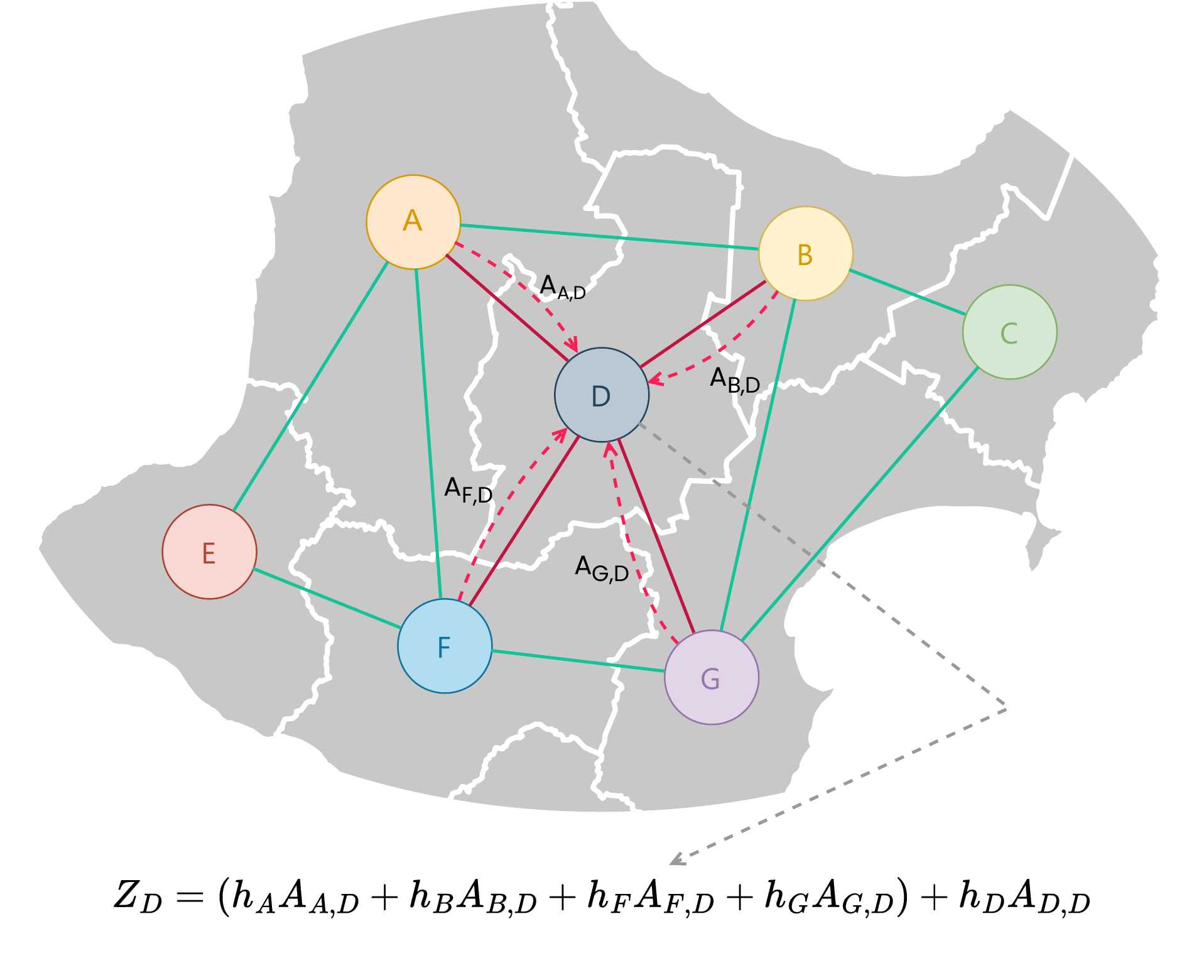}
  \caption{The message-passing mechanism on an example district health board graph-based representation}
\end{figure}

\subsection{Temporal architectures}


\subsubsection{Long Short-Term Memory} \label{sec:LSTM}
Long Short-Term Memory (LSTM), first proposed by \cite{10.1162/neco.1997.9.8.1735}, is a special kind of Recurrent Neural Networks that was designed for learning sequential and time-series data. LSTM has been widely applied into many current state-of-the-art Deep Learning models in various aspects of Machine Learning including Natural Language Processing, Speech Recognition and Computer Vision. One successful variant of LSTM is the Gated Recurrent Unit (GRU) introduced by \cite{cho-etal-2014-learning} as a simplification with less computational cost in the context of sequential modeling. The forward pass of an LSTM cell with a forget gate can be described as follows:
\begin{align*}
f_t & = \sigma_g(W_f x_t + U_f h_{t - 1} + b_f) \in (0, 1)^h & \text{is the forget gate's activation vector} \\
i_t & = \sigma_g(W_i x_t + U_i h_{t - 1} + b_i) \in (0, 1)^h & \text{is the input/update gate's activation vector} \\
o_t & = \sigma_g(W_o x_t + U_o h_{t - 1} + b_o) \in (0, 1)^h & \text{is the output gate's activation vector} \\
\tilde{c}_t & = \sigma_c(W_c x_t + U_c h_{t - 1} + b_c) \in (-1, 1)^h & \text{is the cell input activation vector}\\
c_t & = f_t \odot c_{t - 1} + i_t \odot \tilde{c}_t \in \mathbb{R}^h & \text{is the cell state vector}\\
h_t & = o_t \odot \sigma_h(c_t) \in (-1, 1)^h & \text{is the hidden state (output) vector of the LSTM unit}
\end{align*}
where the initial values are $c_0 = 0$ and $h_0 = 0$; the subscript $t$ indexes the time step; $d$ and $h$ refer to the number of input features and number of hidden units, respectively; $x_t \in \mathbb{R}^d$ is the input vector to the LSTM unit; $W \in \mathbb{R}^{h \times d}$, $U \in \mathbb{R}^{h \times h}$, and $b \in \mathbb{R}^h$ are learnable weight matrices and bias vector; the operator $\odot$ denotes the element-wise (i.e. Hadamard) product; $\sigma_g(\cdot)$ is the sigmoid function; $\sigma_c(\cdot)$ and $\sigma_h(\cdot)$ are the hyperbolic tangent function. \\


\subsubsection{Transformers} \label{sec:Transformers}
Recently, Transformers have achieved superior performances in various deep learning tasks \cite{devlin-etal-2019-bert,dosovitskiy2021an,NIPS2017_3f5ee243}. Among multiple advantages of Transformers, the ability to capture long-range dependencies and interactions is particularly important for time series modeling. The backbone of Transformers is the self-attention mechanism\cite{NIPS2017_3f5ee243}, also called scaled dot-product attention or softmax attention. Self-attention transforms the input sequence $X = [x_1, .., x_L]^T \in \mathbb{R}^{L \times d}$ of length $L$ into the output sequence $H = [h_1, .., h_L]^T \in \mathbb{R}^{L \times h}$ in the following two steps:
\begin{enumerate}
\item The input sequence $X$ is projected into the query matrix $Q = [q_1, .., q_L]^T$, the key matrix $K = [k_1, .., k_L]^T$ and the value matrix $V = [v_1, .., v_L]^T$ via three linear transformations:
$$Q = XW_Q^T, \ \ \ \ K = XW_K^T, \ \ \ \ V = XW_V^T,$$
where $W_Q, W_K, W_V \in \mathbb{R}^{h \times d}$ are learnable weight matrices.
\item The output sequence $H$ is then computed as follows:
\begin{equation}
H = \text{Attention}(Q, K, V) = \text{softmax}\bigg(\frac{QK^T}{\sqrt{h}}\bigg)V = AV,
\label{eq:self-attention}
\end{equation}
where the softmax function is applied to each row of the matrix $QK^T$, and $A \in \mathbb{R}^{L \times L}$ is the attention matrix of $a_{ij}$ attention scores. Equation \ref{eq:self-attention} can be rewritten as:
$$h_i = \sum_{j = 1}^L \text{softmax}\bigg(\frac{q_i^Tk_j}{\sqrt{h}}\bigg) = \sum_{j = 1}^L a_{ij}v_j.$$
\end{enumerate}
Each output sequence $H$ forms an attention head. Let $n$ be the number of heads and $W_O \in \mathbb{R}^{nh \times nh}$ be the projection matrix for the output. In multi-head attention, multiple heads are concatenated to compute the final output defined as follows:
$$\text{Multihead}(\{Q, K, V\}_{i = 1}^n) = \text{Concat}(H_1, .., H_n)W_O.$$


\subsection{Graph Neural Networks}

\subsubsection{Graph construction}

We process the input disease data as graphs, a form of non-Euclidean irregular data that is \textit{permutation invariant} in nature (i.e., changing the ordering of the nodes in a graph does not change the data that the graph represents). To represent the New Zealand pandemic data as graphs, the entirety of the country is formatted as a single graph $G = (V, E)$, where $n = |V|$ is the number of nodes, and each node represents a single district health board in New Zealand. We create a series of graphs $G^{(1)}, G^{(2)},...,G^{(T)}$ corresponding to each day in the case dataset of New Zealand, where the current day $t$ is within the available day case data for every district health board. The topology (i.e., connecting edges and adjacency matrix) of the graphs remains constant over all time steps. The adjacency matrix $\textbf{A}$ represents the connection between edges in the disease graph; we constructed the connections between nodes based on geographical adjacency between any two district health boards. Between any two district health boards $u$ and $v$, the edge $(u, v)$ from $u$ to $v$ is $A_{u,v} = 2$ if two district health boards share any border length, and $A_{u,v}=1$ otherwise. For each node or district health board, we denote the features, or the number of cases in the last $d$ days in the region $u$, as the vector $\textbf{x}^{(t)}_u = (c_u^{(t-d)},...,c_u^{(t)})^\top \in \mathbb{R}^d$. The number of cases over multiple previous days is used to account for irregular case reporting and the length of the incubation period.

\subsubsection{Message-passing neural networks} \label{sec:MPNN}

We model the spatial and geographical spread of COVID-19 in New Zealand using a well-known family of GNNs known as message-passing neural networks (MPNNs)\cite{pmlr-v70-gilmer17a}. Vector messages are exchanged between nodes and updated using neural networks; intuitively, the model takes into account the interaction between neighbors in the network to model the spread of the disease. District health boards with shared borders are more connected and see more traffic between them compared to two boards in remote regions. The goal of the model is to generate a vectorized representation for each node in the disease graph. 

With hidden embedding $h_u^{(k)}$ representing each node/district health board $u \in V$, we define the GNN message-passing mechanism based on the Graph Convolutional Network\cite{kipf2017semisupervised} as
\begin{equation}
    h^{(k)}_u = \sigma \left( W^{(k)} \sum_{v \in \mathcal{N}(u) \cup \{u\}} \frac{h_v}{\sqrt{|\mathcal{N}(u)||\mathcal{N}(u)|}} \right)
\label{eq:single-MPNN}
\end{equation}
where $W^{(k)}$ is the trainable parameter matrix of layer $k$, and $\sigma$ is an element-wise nonlinearity (i.e., ReLU or tanh function). $\mathcal{N}(u)$ denotes the set of neighboring nodes to node $u$; in this case, the set represents all district health boards adjacent to board $u$. Note that the term $\frac{h_v}{\sqrt{|\mathcal{N}(u)||\mathcal{N}(u)|}}$ represents neighborhood normalization based on the degrees of nodes (i.e., number of shared borders) for the purpose of increasing computational stability. The number of message-passing layers $k$ represents the number of aggregation operations, as one layer integrates each node with the information from its adjacent neighbors, while two layers add neighbors two steps away from the target node, etc. 

Across multiple layers and to account for the vectorization of the node embeddings, we define the neighborhood aggregation scheme as
\begin{equation}
    H^{(k)} = \sigma(\tilde{A}H^{(k-1)}W^{(k)})
\label{eq:matrix-MPNN}
\end{equation}
where $H^{(k-1)}$ is a matrix containing the generated node embeddings from the previous layer, $H^{(k)} = (h^{(k)}_1,h^{(k)}_2,...,h^{(k)}_n)^\top$ denotes the matrix arrangement of the node embeddings of all nodes in the graph ($H^{(0)} = X$), and $\tilde{A}$ denotes the aforementioned normalized graph Laplacian. Note that Equation \ref{eq:matrix-MPNN} represents the matrix vectorization of the MPNN described in Equation \ref{eq:single-MPNN}, and thus the two equations are equivalent representations of the same MPNN. For the sake of brevity, the time index is omitted from both equations; the model is in fact applied to all input graphs $G^{(1)}, G^{(2)},...,G^{(T)}$ in the time series separately. Since the connectivity and adjacency of the disease graphs are constant over time, the matrix $\tilde{A}$ is shared across all temporal graphs alongside the weight matrices $W^{(1)},...,W^{(K)}$ for $K$ message-passing layers, while the node embeddings $H^{0},...,H^{K}$ are unique for each disease day graph in the time series.

\begin{figure}[htbp]
  \centering
  \includegraphics[scale=0.8]{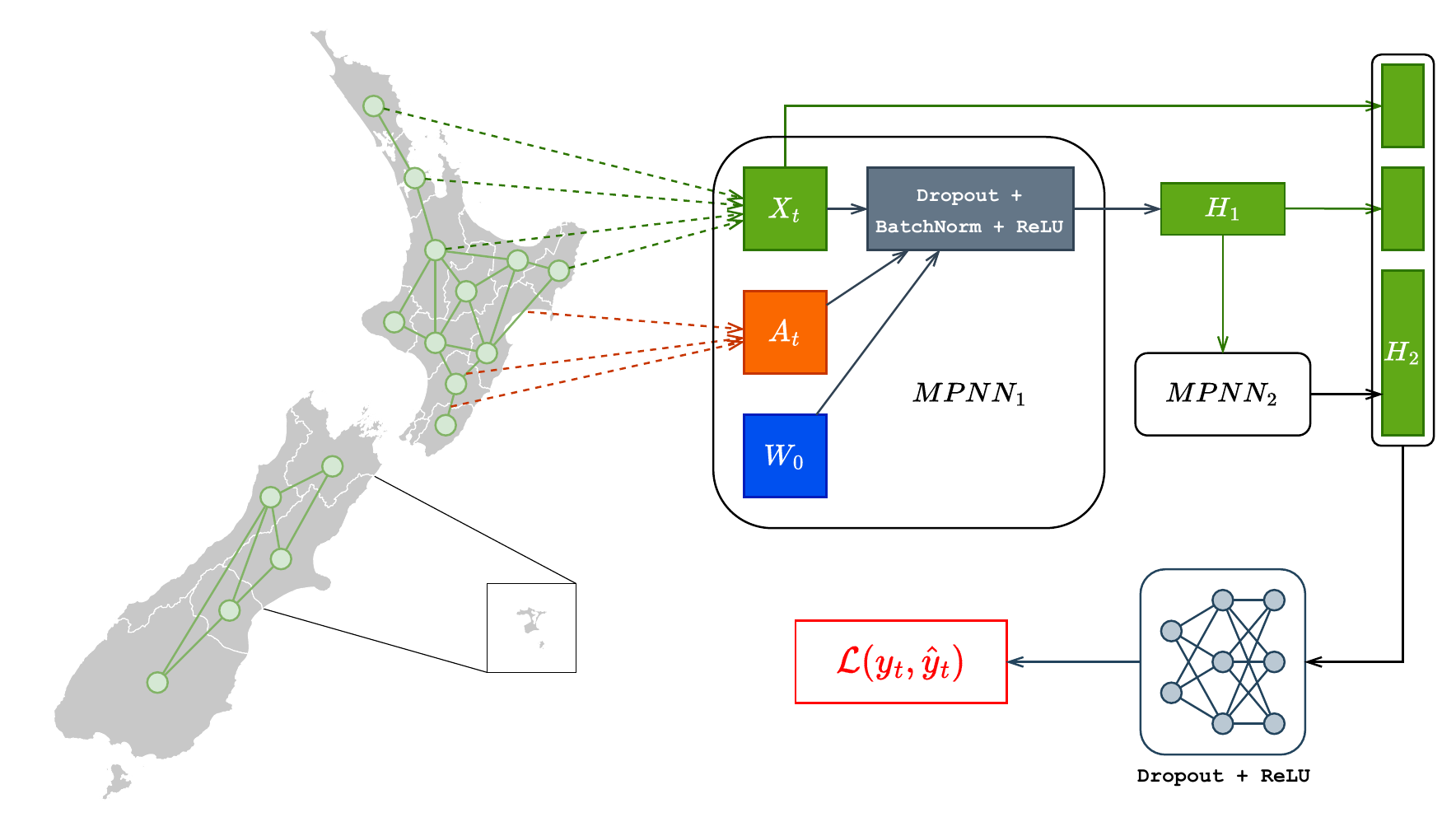}
  \caption{Overview of the proposed Message-Passing Neural Network architecture on the graph representation of New Zealand. Note that dotted green arrows represent the extraction of historical case counts as node features, and dotted orange arrows represent the geospatial location between two regions extracted as edge features.}
\end{figure}

\subsubsection{Multiresolution Graph Neural Networks} \label{sec:MGNN}

In field of graph learning, it is important to build a graph neural network that can capture the multiscale and hierarchical structures of graphs. Multiresolution Graph Neural Networks (MGNN) was originally proposed by \cite{Hy_2023} as a graph encoder in the context of graph generation via variational autoencoder, and adopted by \cite{pmlr-v184-hy22a} in combination with a temporal architecture to learn and predict the dynamics of an epidemic or a pandemic. Instead of a fixed coarse-graining process, MGNN introduces a learnable clustering algorithm that iteratively constructs a hierarchy of coarsening graphs, also called multiresolution or multiple levels of resolutions (see Def.~\ref{def:multiresolution}):
\begin{enumerate}
\item Based on the node embeddings, we cluster a graph into multiple partitions. Each partition is coarsened into a single node, and all the edges connecting between two partitions are aggregated into a single edge (see Def.~\ref{def:coarsening}). This process results into a smaller coarsened graph.
\item We continue to apply message passing on the coarsened graph to produce its node embeddings, and then cluster it further. On each resolution, all the node embeddings are pooled into a single graph-level vectorized representation, i.e. latent. The hierarchy of latents allows us to capture both local information (in the lower levels) and global information (in the higher levels) of a graph.
\end{enumerate}

\begin{definition} \label{def:coarsening}
A $k$-cluster partition on a graph $G = (V, E)$ partitions its set of nodes into $k$ disjoint sets $\{V_1, V_2, .., V_k\}$. A coarsening of $G$ is a graph $\tilde{G} = (\tilde{V}, \tilde{E})$ of $k$ nodes in which node $\tilde{v}_i \in \tilde{V}$ corresponds to a induced subgraph of $G$ on $V_i$. The weighted adjacency matrix $\tilde{A} \in \mathbb{N}^{k \times k}$ of $\tilde{G}$ is defined as:
$$
\tilde{A}_{ij} = 
\begin{cases} 
\frac{1}{2} \sum_{u, v \in V_i} A_{uv}, & \mbox{if } i = j, \\ 
\sum_{u \in V_i, v \in V_j} A_{uv}, & \mbox{if } i \neq j, 
\end{cases}
$$
where the diagonal of $\tilde{A}$ denotes the number of edges inside each cluster, while the off-diagonal denotes the number of edges between two clusters.
\end{definition}

\begin{definition} \label{def:multiresolution}
An $L$-level of resolutions, i.e. multiresolution, of a graph $G$ is a series of $L$ graphs $\tilde{G}_1, .., \tilde{G}_L$ in which: \textbf{(i)} $\tilde{G}_L$ is $G$ itself; and \textbf{(ii)} For $1 \leq \ell \leq L - 1$,~ $\tilde{G}_\ell$ is a coarsening graph of $\tilde{G}_{\ell + 1}$ as defined in Def.~\ref{def:coarsening}. The number of nodes in $\tilde{G}_{\ell}$ is equal to the number of clusters in $\tilde{G}_{\ell + 1}$. The top level coarsening $\tilde{G}_1$ is a graph consisting of a single node. 
\end{definition}

\noindent
The key innovation of MGNN is how the model can learn to cluster graph $\tilde{G}_{\ell + 1}$ into $\tilde{G}_\ell$ in a data-driven manner. Without the loss of generality, we suppose that the number of nodes in $\tilde{G}_\ell$ is $K$, i.e. $|\tilde{V}_\ell| = K$, meaning that we cluster $\tilde{G}_{\ell + 1}$ into $K$ partitions. First, we employ a GNN to produce a $K$-channel node embedding for each node of $\tilde{G}_{\ell + 1}$. Then, we apply a softmax over the node embedding to compute the probability of assigning each node to one of the $K$ clusters. However, we want each node to be in a single cluster, i.e. hard clustering, thus we employ the Gumbel-max trick \cite{Gumbel1954,NIPS2014_309fee4e,jang2017categorical} to sample/select the cluster based on the assignment probability while maintaining differentiability for back-propagation. This results into an assignment matrix $P \in \{0, 1\}^{|\tilde{V}_{\ell + 1}| \times K}$. The adjacency matrix of $\tilde{G}_\ell$ can be computed as $\tilde{A}_\ell = P^T \tilde{A}_{\ell + 1} P$. We repeat this clustering process iteratively in order to build multiple resolutions of coarsening graphs.

\subsubsection{Spatio-temporal graph neural networks} \label{sec:tgnn}

In this section, we build our spatio-temporal GNNs by combining all the previously defined modules. Suppose that we are given a historical data of $T$ timesteps which can be modeled by $T$ input graphs $G^{(1)}, G^{(2)}, .., G^{(T)}$. The simplest combination is MPNN+LSTM in which we employ MPNN (see Section~\ref{sec:MPNN}) to encode each $G^{(t)}$ into a graph-level vectorized representation and then feed it into an LSTM backbone (see Section~\ref{sec:LSTM}). Furthermore, we want to capture the multiscale information, i.e. local to global, that is essential in modeling the long-range spatial and temporal dependencies. Thus, instead of MPNN, we apply MGNN (see Section~\ref{sec:MGNN}) to construct a hierarchy of latents (i.e. each latent is a graph-level representation for a resolution) for each graph $G^{(t)}$. At the $t$-th timestep, a Transformer (see Section~\ref{sec:Transformers}) is applied to encode the hierarchy of latents into a single vector that will be fed further into a temporal architecture. Finally, another Transformer is used, instead of LSTM, as the temporal backbone. We call this novel architecture as Attention-based Multiresolution Graph Neural Networks or ATMGNN.

\subsection{Data preprocessing}
\label{sec:DP}

Our dataset primarily focuses on COVID-19 in New Zealand. The number of cases in different district health boards or regions of New Zealand was gathered from government official open data and later reprocessed into the format suitable for input into the forecasting models. The reprocessed data is available on our GitHub page for the project \url{https://github.com/HySonLab/pandemic_tgnn}. 

\paragraph{New Zealand daily new cases with graphs}

Official data originally obtained is in tabular form with information regarding the sex, age group, district health board location, case status and travel of each COVID-19 infected patient. All cases are filtered so that only cases in 2022 and cases that are confirmed are included in the dataset. For each district health board, on each day, all confirmed cases regardless of sex or age group are aggregated and counted toward the daily new cases count. From the geographical map of the district health boards (Figure \ref{fig:DHB_map}), an adjacency matrix that represents the topology of the disease graph is generated by connecting each board to itself with a unit weighted edge, and each board to every other board that shares any part of its border with edges weighted as 2. Original data is imported and transformed using the Python packages Pandas and NumPy\cite{mckinney-proc-scipy-2010, harris2020array}, while disease graphs are built with the included code and the NetworkX\cite{SciPyProceedings_11} package. All data that was preprocessed and converted to graph form between March \nth{4}, 2022 and November \nth{4}, 2022 is available on GitHub.

\paragraph{New Zealand economic features}

Official categorical GDP data is obtained from NZStats\cite{nzstats}, with the original data containing GDP information in terms of NZ dollars for each predefined administrative region and for every 22 available economic industry categories. At the time of collection, only GDP information until the end of 2020 is finalized and available for all applicable industries. Thus, based on the assumption that categorical GDP from 2020 and 2022 does not change significantly year-to-year, GDP data from 2020 is incorporated as additional features into all time steps of the forecasting model. Due to differences between the administrative region map and the district health board map, all regions between the two maps are matched appropriately, with the merged GDP being the sum or the average between matched regions accordingly. The raw GDP number of each industry/category of each region is concatenated to a common vector of that region without labels to be added to the inputs as a single feature vector. All economic feature vectors are normalized (via mean and standard deviation) to allow the models to learn properly and mitigate exploding/vanishing gradients.

\bibliography{sample}

\clearpage
\section{Appendix}

\begin{table*}[hbt!]
    \centering
    \def\arraystretch{1.1}
    \resizebox{\textwidth}{!}{ 
        \begin{tabular}{|l|rrr|rrr|rrr|rrr|} \hline
        \multirow{2}{*}{Model} & \multicolumn{3}{c|}{Next 3 Days} & \multicolumn{3}{c|}{Next 7 Days} & \multicolumn{3}{c|}{Next 14 Days} & \multicolumn{3}{c|}{Next 21 Days} \\ \cline{2-13}
        & MAE & RMSE & $\text{R}^2$ & MAE & RMSE & $\text{R}^2$ & MAE & RMSE & $\text{R}^2$ & MAE & RMSE & $\text{R}^2$ \\
        \Xhline{3\arrayrulewidth}
        AVG & 7.65 & 14.41 & -462.73 & 7.55 & 14.22 & -591.09 & 7.92 & 15.20 & -872.30 & 8.49 & 16.56 & -1187.11 \\ 
        AVG\_WINDOW & \textbf{5.24} & \textbf{9.47} & -102.43 & 5.69 & 10.14 & -215.84 & 7.88 & 14.55 & -666.37 & 10.56 & 19.55 & -2235.75 \\ 
        LAST\_DAY & 7.29 & 13.94 & -140.97 & 5.05 & \textbf{9.84} & -181.85 & 7.12 & 14.01 & -970.65 & 10.05 & 19.45 & -6024.04 \\ 
        \hline
        LIN\_REG & 9.89 & 19.71 & -0.55 & 12.59 & 24.65 & -3.63 & 18.57 & 35.83 & -21.64 & 26.00 & 49.31 & -101.23 \\ 
        GP\_REG & 8.20 & 17.55 & -0.22 & 7.51 & 13.45 & -0.10 & 8.97 & 18.67 & -5.34 & 9.79 & 21.43 & -18.70 \\ 
        RAND\_FOREST & 6.91 & 16.15 & -0.04 & 5.01 & 10.94 & 0.09 & 7.13 & 15.60 & -3.29 & 9.76 & 20.83 & -17.25 \\ 
        XGBOOST & 7.80 & 18.15 & -0.31 & \textbf{5.00} & 12.56 & -0.20 & \textbf{7.03} & 17.15 & -4.19 & 9.71 & 22.87 & -20.98 \\ 
        \hline
        PROPHET & 11.12 & 42.21 & -1.41 & 13.86 & 44.01 & -2.13 & 21.25 & 40.86 & -3.10 & 27.88 & 39.83 & -8.04 \\ 
        ARIMA & 9.09 & 19.95 & \textbf{0.46} & 9.08 & 20.54 & \textbf{0.32} & 8.78 & 16.03 & \textbf{0.37} & \textbf{8.13} & \textbf{13.37} & \textbf{-0.03} \\ 
        LSTM & 7.95 & 14.98 & -590.28 & 6.12 & 11.62 & -41.90 & 7.89 & 14.42 & -29.37 & 8.93 & 16.34 & -35.69 \\
        \hline
        \textbf{MPNN}  & 6.41 & 12.14 & -28.73 & 5.61 & 10.70 & -41.77 & 8.11 & 15.09 & -254.37 & 10.85 & 20.89 & -1358.29 \\ 
        \textbf{MGNN} & 7.04 & 11.83 & -27.92 & 7.44 & 11.66 & -69.09 & 10.17 & 16.27 & -424.27 & 13.21 & 22.09 & -1363.33 \\ 
        \textbf{MPNN+LSTM} & 6.92 & 12.72 & -19.68 & 7.55 & 13.71 & -30.45 & 7.46 & 13.10 & -6.30 & 8.26 & 15.20 & -2.66 \\ 
        \textbf{ATMGNN} & 7.44 & 13.21 & -13.40 & 7.16 & 12.59 & -27.34 & 7.28 & \textbf{12.69} & -9.79 & 8.25 & 14.18 & -1.61 \\
        \hline        \end{tabular}
    }
    \caption{France: Performance of all experimental model evaluated based on the metrics specified in Section 2.5}
    \label{tab:FR_results}
\end{table*}

\begin{table*}[hbt!]
    \centering
    \def\arraystretch{1.1}
    \resizebox{\textwidth}{!}{ 
        \begin{tabular}{|l|rrr|rrr|rrr|rrr|} \hline
        \multirow{2}{*}{Model} & \multicolumn{3}{c|}{Next 3 Days} & \multicolumn{3}{c|}{Next 7 Days} & \multicolumn{3}{c|}{Next 14 Days} & \multicolumn{3}{c|}{Next 21 Days} \\ \cline{2-13}
        & MAE & RMSE & $\text{R}^2$ & MAE & RMSE & $\text{R}^2$ & MAE & RMSE & $\text{R}^2$ & MAE & RMSE & $\text{R}^2$ \\
        \Xhline{3\arrayrulewidth}
        AVG & 48.71 & 111.19 & -9.68 & 52.60 & 122.82 & -12.68 & 60.01 & 149.53 & -20.01 & 68.19 & 178.26 & -31.68 \\ 
        AVG\_WINDOW & \textbf{32.56} & \textbf{59.57} & -0.51 & 40.09 & 79.83 & -3.08 & 53.03 & 121.30 & -12.42 & 63.15 & 159.29 & -25.10 \\ 
        LAST\_DAY & 35.20 & 63.98 & -0.34 & 37.63 & 70.57 & -1.23 & 52.60 & 112.27 & -8.45 & 63.23 & 155.97 & -22.17 \\ 
        \hline
        LIN\_REG & 50.73 & 104.21 & 0.11 & 62.34 & 126.85 & -0.39 & 87.19 & 190.79 & -5.48 & 120.59 & 267.01 & -29.84 \\ 
        GP\_REG & 53.22 & 121.87 & -0.22 & 51.28 & 119.04 & -0.23 & 43.12 & 86.49 & -0.33 & \textbf{31.58} & \textbf{57.52} & -0.43 \\ 
        RAND\_FOREST & 33.27 & 65.77 & \textbf{0.64} & 37.05 & 74.41 & \textbf{0.52} & 51.72 & 117.37 & -1.45 & 61.38 & 155.99 & -9.53 \\ 
        XGBOOST & 35.41 & 69.85 & 0.60 & 38.18 & 76.67 & 0.49 & 52.58 & 117.90 & -1.47 & 62.70 & 159.38 & -9.99 \\
        \hline
        PROPHET & 60.60 & 351.20 & -2.49 & 75.86 & 320.15 & -2.88 & 114.87 & 192.33 & -0.91 & 149.51 & 167.25 & -1.09 \\ 
        ARIMA & 41.89 & 112.37 & \textbf{0.64} & 40.54 & 75.95 & 0.78 & 48.46 & 82.08 & 0.65 & 56.45 & 119.47 & -0.18 \\ 
        LSTM & 43.39 & 84.78 & -3.69 & 44.25 & 85.54 & -3.48 & 36.50 & \textbf{66.01} & 0.33 & 35.74 & 60.87 & \textbf{0.76} \\
        \hline
        \textbf{MPNN}  & 33.26 & 65.51 & 0.24 & 39.91 & 82.70 & -0.70 & 50.42 & 106.45 & -4.64 & 61.95 & 142.19 & -9.40 \\ 
        \textbf{MGNN} & 34.71 & 66.16 & 0.26 & 42.35 & 84.48 & -0.68 & 54.31 & 109.22 & -5.09 & 67.11 & 148.25 & -10.40 \\ 
        \textbf{MPNN+LSTM} & 34.60 & 69.55 & 0.37 & \textbf{35.03} & \textbf{67.52} & 0.45 & 37.95 & 81.06 & 0.56 & 39.68 & 82.58 & 0.72 \\ 
        \textbf{ATMGNN} & 34.08 & 69.37 & 0.40 & 35.88 & 72.06 & 0.40 & \textbf{34.73} & 69.56 & \textbf{0.70} & 38.13 & 82.72 & 0.71 \\
        \hline        \end{tabular}
    }
    \caption{Spain: Performance of all experimental model evaluated based on the metrics specified in Section 2.5}
    \label{tab:ES_results}
\end{table*}

\begin{table*}[hbt!]
    \centering
    \def\arraystretch{1.1}
    \resizebox{\textwidth}{!}{ 
        \begin{tabular}{|l|rrr|rrr|rrr|rrr|} \hline
        \multirow{2}{*}{Model} & \multicolumn{3}{c|}{Next 3 Days} & \multicolumn{3}{c|}{Next 7 Days} & \multicolumn{3}{c|}{Next 14 Days} & \multicolumn{3}{c|}{Next 21 Days} \\ \cline{2-13}
        & MAE & RMSE & $\text{R}^2$ & MAE & RMSE & $\text{R}^2$ & MAE & RMSE & $\text{R}^2$ & MAE & RMSE & $\text{R}^2$ \\
        \Xhline{3\arrayrulewidth}
        MPNN+LSTM & 128.79 & 139.09 & -3.65 & 149.44 & 160.53 & -2.73 & 174.22 & 185.44 & -2.81 & 203.13 & 214.37 & -3.17 \\ 
        ATMGNN & 132.83 & 143.38 & -4.04 & 150.52 & 161.47 & -2.77 & 178.19 & 189.93 & -2.88 & 208.49 & 220.03 & -3.20 \\        \hline 
        \textbf{MPNN+LSTM (ECON)} & 128.17 & 139.26 & -3.22 & 150.13 & 161.30 & -3.00 & 165.94 & 178.23 & -2.62 & 199.19 & 211.00 & -3.17 \\ 
        \textbf{ATMGNN (ECON)} & 131.59 & 142.21 & -4.23 & 154.19 & 165.13 & -3.25 & 173.04 & 184.48 & -2.71 & 203.36 & 214.87 & -3.16 \\
        \hline
        \end{tabular}
    }
    \caption{New Zealand: Limited DHB map results for graph-based models with and without economic features}
    \label{tab:NZ_results_econ_lim}
\end{table*}

\begin{table*}[hbt!]
    \centering
    \def\arraystretch{1.1}
    \resizebox{\textwidth}{!}{ 
        \begin{tabular}{|l|rrr|rrr|rrr|rrr|} \hline
        \multirow{2}{*}{Model} & \multicolumn{3}{c|}{Next 3 Days} & \multicolumn{3}{c|}{Next 7 Days} & \multicolumn{3}{c|}{Next 14 Days} & \multicolumn{3}{c|}{Next 21 Days} \\ \cline{2-13}
        & MAE & RMSE & $\text{R}^2$ & MAE & RMSE & $\text{R}^2$ & MAE & RMSE & $\text{R}^2$ & MAE & RMSE & $\text{R}^2$ \\
        \Xhline{3\arrayrulewidth}
        MPNN+LSTM (MO) & 86.85 & 119.97 & 0.84 & 99.89 & 137.90 & 0.81 & 114.84 & 158.12 & 0.75 & 128.79 & 177.60 & 0.70 \\ 
        ATMGNN (MO) & 85.67 & 118.27 & 0.84 & 100.91 & 138.11 & 0.81 & 114.43 & 157.87 & 0.76 & 129.89 & 178.97 & 0.70 \\        \hline 
        \textbf{MPNN+LSTM (MO+ET)} & 109.82 & 150.23 & 0.76 & 121.45 & 166.33 & 0.73 & 124.55 & 171.37 & 0.71 & 136.98 & 188.01 & 0.66 \\ 
        \textbf{ATMGNN (MO+ET)} & 104.44 & 143.88 & 0.76 & 113.65 & 155.83 & 0.76 & 128.81 & 177.27 & 0.70 & 133.34 & 183.91 & 0.68 \\
        \hline
        \end{tabular}

    }
    \caption{New Zealand: Multiple outputs demographical graph models results, with and without explicit training weighting. MO denotes the multiple output graph models, and ET denotes custom age group weighting during graph model training.  }
    \label{tab:NZ_results_mo}
\end{table*}

\end{document}